\documentclass[10pt,twocolumn]{article}

\usepackage[utf8]{inputenc}
\usepackage[T1]{fontenc}
\usepackage{times}
\usepackage{microtype}
\usepackage{graphicx}
\usepackage{booktabs}
\usepackage{amsmath,amssymb}
\usepackage{hyperref}
\usepackage{xcolor}
\usepackage{natbib}
\usepackage{caption}
\usepackage{tabularx}
\usepackage{float}
\usepackage{authblk}
\usepackage[margin=1in]{geometry}

\hypersetup{
  colorlinks=true,
  linkcolor=blue!60!black,
  citecolor=blue!60!black,
  urlcolor=blue!60!black,
}

\title{\Large \bf Extending TotalSegmentator: Predicting Patient and Acquisition Characteristics from CT and MR Images
}

\author[1]{Jakob Wasserthal\thanks{Corresponding author: \texttt{jakob.wasserthal@usb.ch}}}
\author[1]{Joshy Cyriac}
\author[1]{Michael Bach}
\author[2,3]{Kimia Mozahheb Yousefi}
\author[4,5]{Minh-Son To}
\author[6,7]{Máté Sik}
\author[8]{Cédric Hémon}
\author[9]{Thomas Weikert}
\author[10]{Marwan Abbas}
\author[1]{Martin Segeroth}

\affil[1]{Clinic of Radiology and Nuclear Medicine, University Hospital Basel, Basel, Switzerland}
\affil[2]{Antimicrobial Resistance Research Center, Institute of Immunology and Infectious Diseases, Iran University of Medical Sciences, Tehran, Iran}
\affil[3]{Faculty of Medicine, Iran University of Medical Sciences, Tehran, Iran}
\affil[4]{Flinders Health and Medical Research Institute, Flinders University, Bedford Park, Australia}
\affil[5]{South Australia Medical Imaging, Flinders Medical Centre, Bedford Park, Australia}
\affil[6]{Division of Radiology and Imaging Science, Department of Medical Imaging, Faculty of Medicine, University of Debrecen, Debrecen, Hungary}
\affil[7]{Doctoral School of Medical Sciences, University of Debrecen, Debrecen, Hungary}
\affil[8]{Univ. Rennes, CLCC Eugène Marquis, INSERM, LTSI - UMR 1099, Rennes, France}
\affil[9]{Zentrum für Bilddiagnostik, Basel, Switzerland}
\affil[10]{Mayo Clinic, Rochester, USA}
\date{}

\begin{document}
\maketitle
\thispagestyle{empty}

\begin{abstract}
\textbf{Background:} Patient details (weight, height, age, sex) and acquisition metadata (e.g. reconstruction kernel, noise, contrast phase, field of view) are important for clinical decisions, image quality control, and automated research pipelines, but may be missing or unreliable in imaging archives.

\textbf{Purpose:} To develop and evaluate a fast and robust open-source model that predicts patient and acquisition characteristics directly from CT and MR images.

\textbf{Materials and Methods:} In this retrospective study, separate 3D ResNet-10 ensembles for CT and MR were trained with 57{,}291 and 43{,}200 clinical examinations acquired from January 2011 to August 2025. Both models jointly predicted weight, height, age, sex, contrast presence, cranial and caudal vertebral coverage, and image noise; the CT model additionally predicted scanner manufacturer, tube voltage, tube current, convolution kernel, and post-injection time, whereas the MR model predicted sequence class. Performance was evaluated on internal CT ($n=501$) and MR ($n=636$) test sets and an external CT dataset ($n=54$). Regression performance was summarized with mean absolute error (MAE), classification with F1 score, and paired models with Wilcoxon signed-rank or exact McNemar tests with Holm correction.

\textbf{Results:} Internal CT MAEs were 3.90\,kg, 3.68\,cm, and 4.42 years for weight, height, and age, with sex F1 of 0.990; corresponding MR results were 4.34\,kg, 4.62\,cm, 7.13 years, and 0.970. The CNN outperformed a segmentation-derived XGBoost baseline for all four core targets in both modalities (adjusted $P\leq.042$). Additional-target performance included F1 scores of 0.963 for CT contrast, 0.953 for MR sequence, and 0.823 for MR contrast. On external full-coverage CT, MAEs were 4.45\,kg, 4.05\,cm, and 5.17 years, with sex F1 of 0.971. CPU inference required 20 seconds for CT and 12 seconds for MR, supporting CPU-only deployment.

\textbf{Conclusion:} One 3D multitask model per modality can rapidly recover a broad set of patient and acquisition characteristics from heterogeneous CT and MR examinations, supporting metadata quality control and automated image-processing pipelines. The models are available in TotalSegmentator (\url{https://github.com/wasserth/TotalSegmentator}).
\end{abstract}

\section{Introduction}
\label{sec:introduction}

Patient weight, height, age, and sex affect contrast administration, drug dosing, radiation-dose optimization, and risk stratification~\citep{bae2010,griggs2012,regitz2012}. Weight and height also define body mass index (BMI) and body surface area (BSA)~\citep{flegal2013,mosteller1987}. Yet these values may be missing, outdated, or estimated in routine metadata~\citep{hall2005,lin2009}.

Acquisition characteristics are similarly important. Tube voltage, tube current, reconstruction kernel, and noise affect CT dose, image appearance, and quantitative reproducibility~\citep{kaza2014,christianson2015,choe2019}. Contrast presence and post-injection timing determine enhancement phase, while MR sequence labels determine which downstream algorithms and quantitative measurements are appropriate~\citep{muhamedrahimov2022,cluceru2023}. Cranial and caudal vertebral levels provide an image-based description of anatomical coverage that can route examinations to compatible pipelines and identify excessive or incomplete coverage~\citep{zhang2015,lessmann2019}. Automated recovery of these properties is therefore relevant to archive quality control, cohort curation, protocol harmonization, and large-scale image analysis.

Prior work estimated individual body characteristics from CT scouts, localizers, dose metrics, tissue segmentations, or thoracoabdominal images~\citep{ichikawa2021,demircioglu2023,ichikawa2024,schenkl2025,kerber2023}. Other image-based systems classified CT contrast or MR acquisition type~\citep{li2024,cluceru2023}, and multitask prediction of MR acquisition parameters has been demonstrated in breast imaging~\citep{konz2024}. However, a single open-source tool that jointly predicts patient characteristics and a broad modality-specific set of acquisition properties from both CT and MR volumes has not been established.

\begin{figure*}[t]
  \centering
  \IfFileExists{imgs/body_stats_overview_cnn.png}{%
    \includegraphics[width=\textwidth]{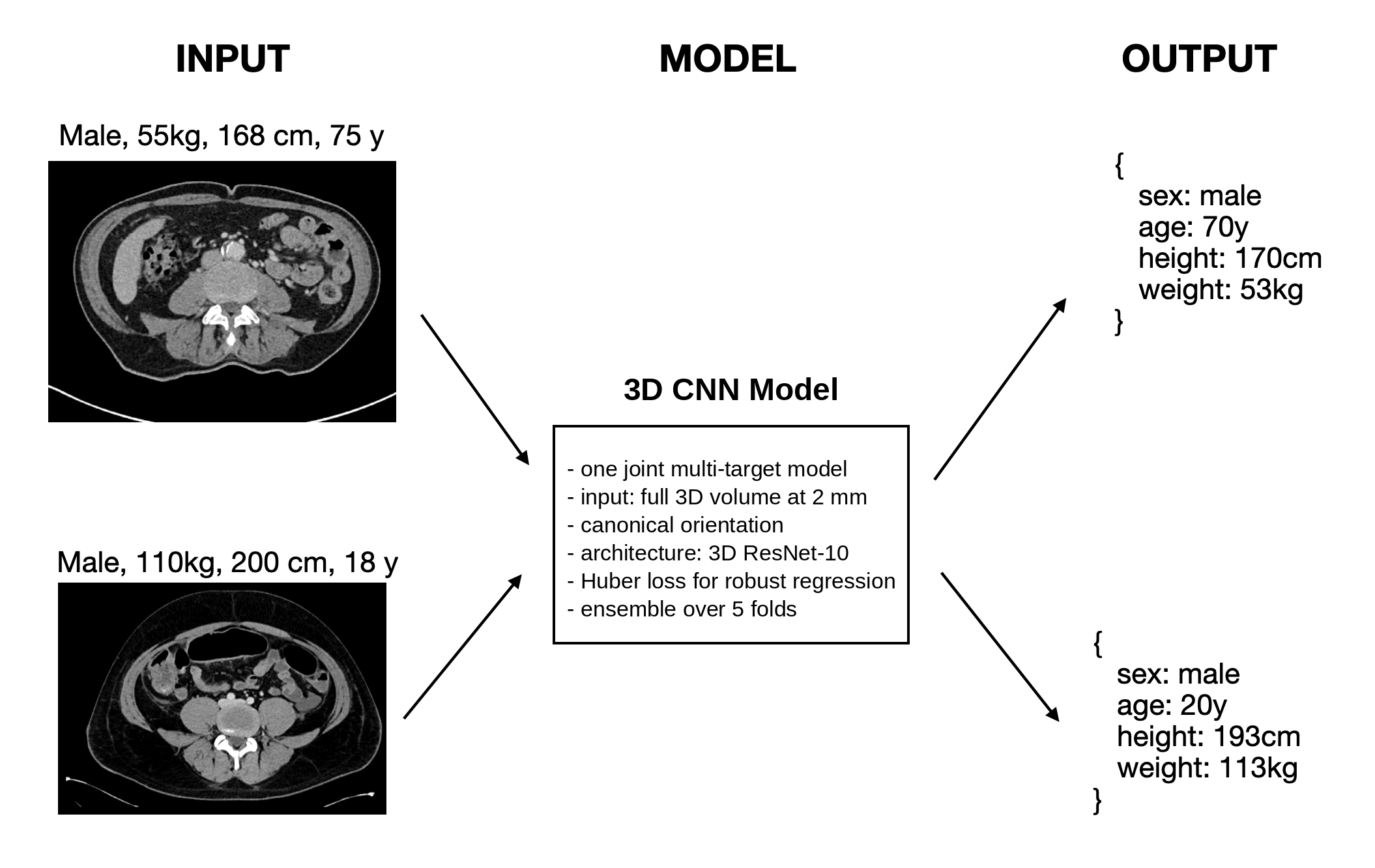}%
  }{%
    \fbox{\parbox{0.92\textwidth}{\centering Missing file: \texttt{imgs/body\_stats\_overview\_cnn.png}}}%
  }
  \caption{Overview of the 3D CNN method. The complete resampled volume is processed by one modality-specific multitask model, and predictions are averaged over five folds. The illustration shows the four core outputs; the same forward pass produces all modality-specific acquisition and quality-control outputs.}
  \label{fig:cnn_overview}
\end{figure*}

The purpose of this study was to develop and evaluate one fast 3D multitask CNN per modality for predicting patient and acquisition characteristics from heterogeneous CT and MR images (Fig.~\ref{fig:cnn_overview}). We compared core outputs with a segmentation-derived XGBoost baseline, evaluated external CT generalization, and tested whether joint prediction reduced accuracy relative to target-specific models.

\section{Materials and Methods}
\label{sec:methods}

\subsection{Study Design and Data}

This retrospective study was approved by the Ethics Committee of Northwest and Central Switzerland (EKNZ BASEC 2023-00446). Clinical examinations acquired between January 2011 and August 2025 were obtained from the institutional picture archiving and communication system; patients who had opted out of research use were excluded. Training required valid DICOM weight (35--160\,kg), height (125--220\,cm), age (12--100 years), and binary sex labels. One series was selected per examination. Torso examinations were prioritized, while highly localized head, joint, or extremity series and modality-specific low-information protocols were deprioritized.

\begin{figure*}[t]
  \centering
  \IfFileExists{imgs/body_stats_data_distribution_all_ext_ct.png}{%
    \includegraphics[width=0.94\textwidth]{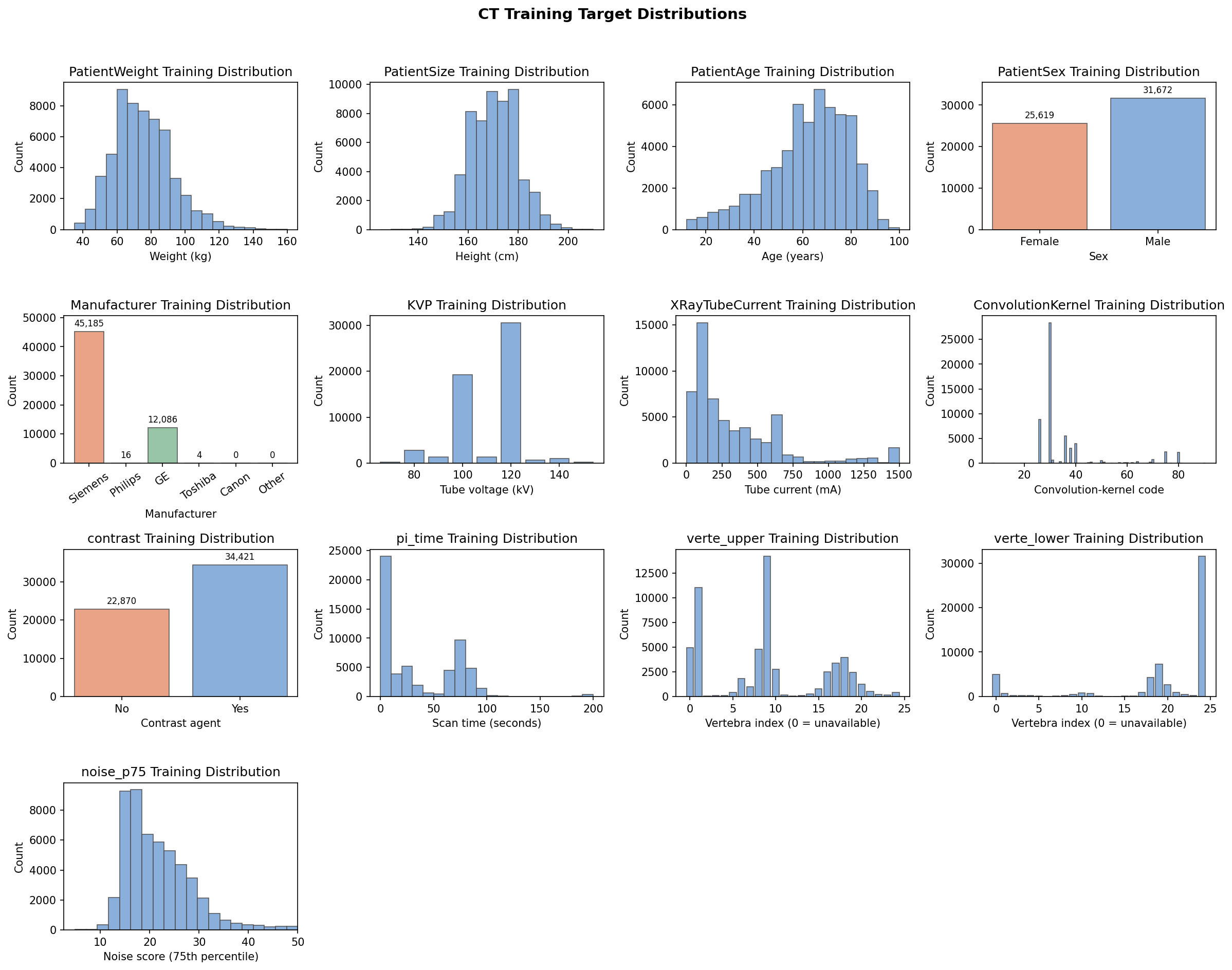}\\[3pt]
    \includegraphics[width=0.94\textwidth]{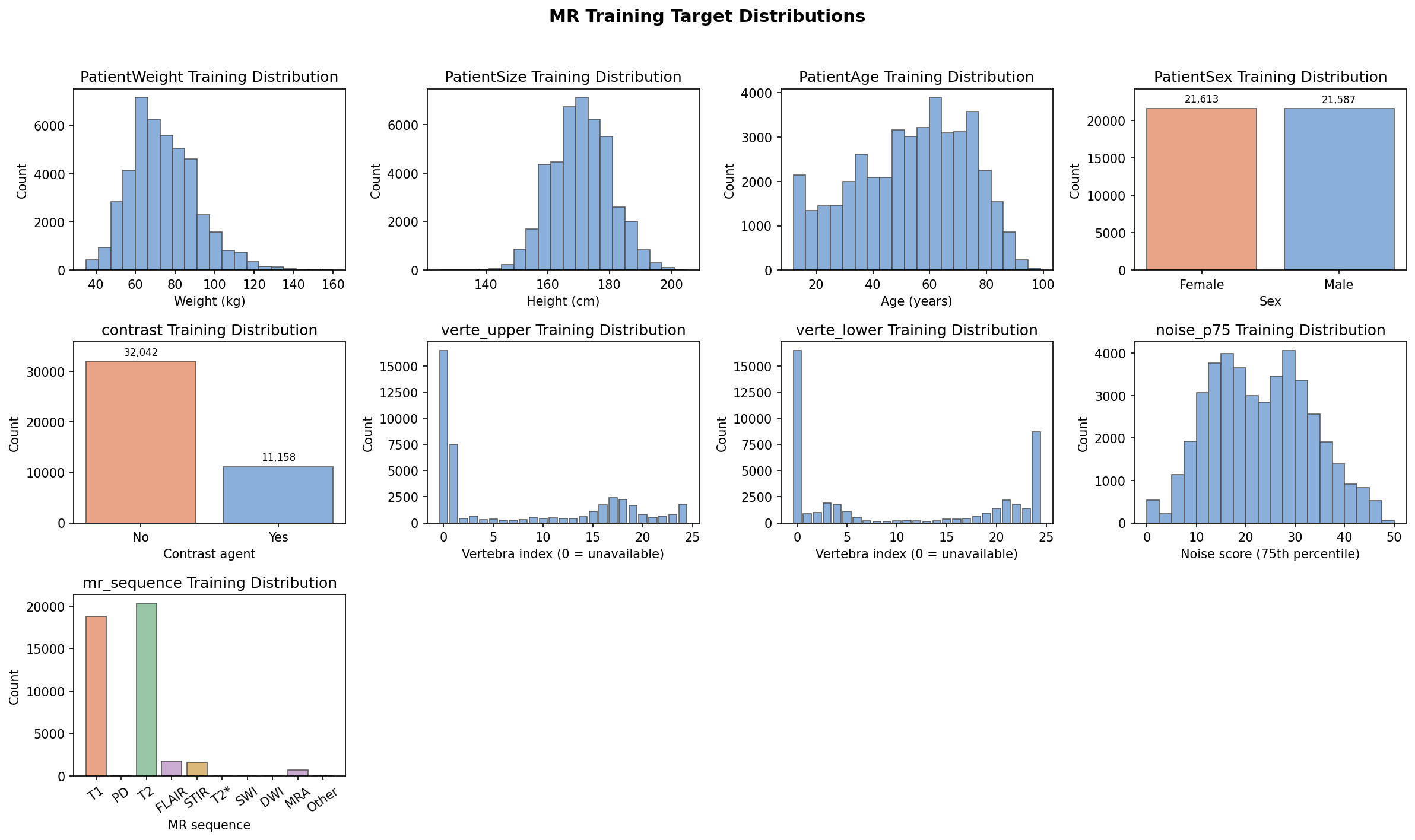}%
  }{%
    \fbox{\parbox{0.92\textwidth}{\centering Missing updated training-distribution figures.}}%
  }
  \caption{Training-label distributions for the CT (top) and MR (bottom) multitask models. Vertebral indices span C1 through L5; zero denotes unavailable coverage labels.}
  \label{fig:data_distribution}
\end{figure*}

The training cohorts contained 57{,}291 CT examinations from 34{,}257 patients and 43{,}200 MR examinations from 29{,}073 patients. Independent internal test sets contained 501 CT and 636 MR examinations with mixed fields of view. External evaluation used 54 examinations from Spine-Mets-CT-SEG~\citep{spinemets}. Full thorax-abdomen-pelvis volumes were additionally cropped to thorax-only and abdomen-pelvis fields of view. Cohort characteristics are shown in Table~\ref{tab:cohorts}, and training-label distributions are shown in Figure~\ref{fig:data_distribution}.

\begin{table*}[t]
  \centering
  \caption{Cohort characteristics. Age is mean $\pm$ standard deviation.}
  \label{tab:cohorts}
  \small
  \begin{tabular}{@{}llrrrc@{}}
    \toprule
    Cohort & Modality & Examinations & Patients & Age (years) & Women \\
    \midrule
    Training & CT & 57{,}291 & 34{,}257 & $62.6 \pm 16.8$ & 25{,}619 (44.7\%) \\
    Training & MR & 43{,}200 & 29{,}073 & $53.4 \pm 20.1$ & 21{,}613 (50.0\%) \\
    Internal test & CT & 501 & 482 & $64.5 \pm 15.1$ & 200 (39.9\%) \\
    Internal test & MR & 636 & 612 & $57.3 \pm 17.8$ & 240 (37.7\%) \\
    External test & CT & 54 & 54 & $64.9 \pm 11.7$ & 19 (35.2\%) \\
    \bottomrule
  \end{tabular}
\end{table*}

\subsection{Prediction Targets and Label Generation}

Both models predicted weight, height, age, and sex. CT targets additionally comprised scanner manufacturer, tube voltage, tube current, a vendor-harmonized convolution-kernel code, intravenous contrast presence, post-injection time, cranial-most and caudal-most visible vertebra, and image-noise score. Reported MR targets additionally comprised contrast presence, both vertebral boundaries, noise score, and MR sequence class (T1, proton density, T2, FLAIR, STIR, T2*, susceptibility-weighted, diffusion-weighted, MR angiography, or other).

Core labels and acquisition parameters were obtained from DICOM metadata. Contrast presence was derived from contrast-agent fields and positive post-injection time. When CT post-injection time was absent or invalid, an existing contrast-phase estimate was used. Vertebral boundaries were the first and last C1--L5 vertebrae with segmented volume greater than 100 voxels. Convolution kernels were mapped to ordinal sharpness codes. Noise ground truth was calculated from local tissue patches as detailed in the Supplementary Material; the model target was the 75th-percentile noise score.

\subsection{Three-dimensional Multitask CNN}

Volumes were reoriented to closest canonical orientation and resampled to 2-mm isotropic spacing. CT intensities were clipped to the training-set 2nd--98th percentiles and normalized with the global training mean and standard deviation; MR volumes were standardized individually. Inputs were center padded or cropped to $240\times240\times240$ voxels for CT and $210\times210\times150$ voxels for MR.

Each modality used a 3D ResNet-10 with one shared linear output layer. All continuous and encoded categorical targets were z standardized using training-fold statistics and optimized jointly with Huber loss ($\beta=1$). Binary outputs were thresholded at 0.5 and multiclass codes were rounded to the nearest valid class at inference. Training used random spatial crops and independent flips along each axis, Adam optimization (learning rate $5\times10^{-4}$), cosine annealing, batch size 16, mixed precision, and 40 epochs. Five examination-level folds were used; the checkpoint with lowest validation loss was retained for each fold. Final predictions were arithmetic means of denormalized outputs from all five models.

\subsection{XGBoost Baseline}

The baseline used TotalSegmentator to segment organs, bones, vertebrae, and tissue compartments~\citep{wasserthal2023,wasserthal2025}. Region volumes and median intensities and vertebral-level subcutaneous fat, torso fat, and skeletal-muscle measurements formed the input to target-specific XGBoost ensembles~\citep{chen2016}. The baseline predicted the four core targets and required the preceding segmentation pipeline.

\subsection{Derived Measures}

BMI and BSA were computed from the predicted weight and height. BMI was calculated as weight in kilograms divided by height in meters squared. BSA was calculated with the Mosteller formula,
\[
\mathrm{BSA} = \sqrt{\frac{\mathrm{height(cm)} \times \mathrm{weight(kg)}}{3600}}.
\]

\subsection{Statistical Analysis}

Regression performance was summarized as mean absolute error (MAE), standard deviation, and interquartile range of absolute errors. Binary F1 used male and contrast-enhanced as positive classes; multiclass targets used micro-F1. Paired regression errors were compared with two-sided Wilcoxon signed-rank tests and sex correctness with exact two-sided McNemar tests. Holm correction was applied across all 28 comparisons. Two-sided adjusted $P<.05$ indicated significance. Runtime and peak system memory were measured for the five-fold CPU ensemble.

\section{Results}
\label{sec:results}

\subsection{Internal Test Sets}

The CNN had lower error than XGBoost for every core regression target in both modalities and higher sex classification performance (Table~\ref{tab:internal}). All eight paired comparisons remained significant after Holm correction (CT adjusted $P\leq.042$; MR adjusted $P\leq.030$). CT CNN MAEs were 3.90\,kg, 3.68\,cm, and 4.42 years for weight, height, and age; MR MAEs were 4.34\,kg, 4.62\,cm, and 7.13 years. Sex F1 was 0.990 for CT and 0.970 for MR.

\begin{table*}[t]
  \centering
  \caption{Core-target performance on internal test sets. Regression values are mean absolute error with standard deviation in parentheses. Sex values are F1 scores with male as the positive class.}
  \label{tab:internal}
  \small
  \begin{tabular}{@{}llrrrrr@{}}
    \toprule
    Modality & Model & $n$ & Weight & Height & Age & Sex \\
    \midrule
    CT & CNN & 501 & 3.90\,kg (4.18) & 3.68\,cm (2.86) & 4.42\,yr (3.50) & 0.990 \\
    CT & XGBoost & 501 & 4.90\,kg (4.76) & 4.29\,cm (3.31) & 6.60\,yr (5.31) & 0.969 \\
    MR & CNN & 636 & 4.34\,kg (4.32) & 4.62\,cm (3.61) & 7.13\,yr (5.95) & 0.970 \\
    MR & XGBoost & 636 & 7.08\,kg (7.60) & 5.21\,cm (4.42) & 11.46\,yr (8.71) & 0.932 \\
    \bottomrule
  \end{tabular}
\end{table*}

\subsection{Additional Targets and Multitask Comparisons}

Additional-target results are shown in Table~\ref{tab:additional}. CT contrast F1 was 0.963, manufacturer micro-F1 was 0.988, post-injection-time MAE was 5.36 seconds, and upper and lower coverage errors were 0.75 and 0.21 vertebral levels. MR sequence micro-F1 was 0.953; MR contrast F1 was 0.823, and vertebral-boundary errors were 2.83 and 2.50 levels.

\begin{table*}[t]
  \centering
  \caption{Performance for additional outputs of the combined models. Values for continuous and ordinal targets are mean absolute error with standard deviation in parentheses; classification values are F1. A dash indicates that the target was not reported for that modality.}
  \label{tab:additional}
  \small
  \begin{tabular}{@{}lcc@{}}
    \toprule
    Target and metric & CT & MR \\
    \midrule
    Manufacturer, micro-F1 & 0.988 & -- \\
    Tube voltage, kV & 5.58 (6.72) & -- \\
    Tube current, mA & 165.84 (223.64) & -- \\
    Convolution-kernel code & 2.29 (2.64) & -- \\
    Contrast presence, F1 & 0.963 & 0.823 \\
    Post-injection time, s & 5.36 (7.24) & -- \\
    Cranial vertebral boundary, levels & 0.75 (1.28) & 2.83 (4.44) \\
    Caudal vertebral boundary, levels & 0.21 (0.45) & 2.50 (4.51) \\
    75th-percentile noise score & 1.54 (1.91) & 2.34 (2.15) \\
    MR sequence, micro-F1 & -- & 0.953 \\
    \bottomrule
  \end{tabular}
\end{table*}

In the fold-matched CT experiment, one combined core model was not significantly different from four separate target-specific models for weight, height, age, or sex (adjusted $P\geq.589$). Adding the extended targets to the five-fold CT model did not significantly change weight, height, or sex performance (adjusted $P\geq.838$), but age MAE increased from 4.02 to 4.42 years (adjusted $P<.001$).

\subsection{External CT Test Set}

On full thorax-abdomen-pelvis CT, CNN MAEs were 4.45\,kg, 4.05\,cm, and 5.17 years, with sex F1 of 0.971 (Table~\ref{tab:external}); no full-coverage CNN--XGBoost comparison was significant after correction. On abdomen-pelvis CT, CNN errors were lower for weight (3.78 vs 7.63\,kg; adjusted $P=.005$) and height (4.53 vs 6.48\,cm; adjusted $P=.045$). On thorax-only CT, CNN age error was lower (5.86 vs 11.53 years; adjusted $P<.001$); the remaining thorax-only comparisons were not significant.

\begin{table*}[t]
  \centering
  \caption{Performance on the external Spine-Mets-CT-SEG CT test set ($n=54$). Regression values are mean absolute error with standard deviation in parentheses. Sex values are F1 scores with male as the positive class.}
  \label{tab:external}
  \small
  \begin{tabular}{@{}llrrrrr@{}}
    \toprule
    Field of view & Model & $n$ & Weight & Height & Age & Sex \\
    \midrule
    Thorax-abdomen-pelvis & CNN & 54 & 4.45\,kg (3.24) & 4.05\,cm (2.82) & 5.17\,yr (3.57) & 0.971 \\
    Thorax-abdomen-pelvis & XGBoost & 54 & 5.26\,kg (5.35) & 4.64\,cm (3.91) & 7.49\,yr (5.39) & 0.971 \\
    Thorax only & CNN & 54 & 6.24\,kg (5.26) & 4.83\,cm (3.81) & 5.86\,yr (4.73) & 0.941 \\
    Thorax only & XGBoost & 54 & 8.63\,kg (6.90) & 6.05\,cm (4.88) & 11.53\,yr (8.18) & 0.909 \\
    Abdomen-pelvis & CNN & 54 & 3.78\,kg (3.39) & 4.53\,cm (3.04) & 5.39\,yr (4.44) & 0.972 \\
    Abdomen-pelvis & XGBoost & 54 & 7.63\,kg (7.61) & 6.48\,cm (5.26) & 8.44\,yr (6.45) & 0.928 \\
    \bottomrule
  \end{tabular}
\end{table*}

\subsection{Runtime}

Five-fold CPU inference required 20 seconds and 3.8\,GB peak RAM for a $512\times512\times807$ CT volume and 12 seconds and 1.7\,GB for a $320\times250\times72$ MR volume. All modality-specific outputs were produced in the same forward passes.

\section{Discussion}
\label{sec:discussion}

This study developed a 3D multitask model to infer patient characteristics and a broad set of acquisition and quality-control properties. The models processed the complete resampled volume, outperformed a segmentation-derived baseline for all core targets, and produced all outputs with short CPU runtimes. A combined core model performed similarly to separate target-specific models, supporting joint inference. Adding extended targets preserved CT weight, height, and sex performance, although the significant increase in age error shows that multitask expansion was not cost-free.

The acquisition outputs address practical archive problems beyond body statistics. Contrast and post-injection timing can help identify phase when series descriptions are inconsistent, as previously demonstrated with dedicated CT systems~\citep{li2024,muhamedrahimov2022}. MR sequence classification enables automated selection of compatible analysis pipelines~\citep{cluceru2023}. Predicted noise, tube settings, and convolution-kernel code can support protocol auditing and quantitative harmonization because these properties alter image quality and radiomic measurements~\citep{kaza2014,christianson2015,choe2019}. Vertebral boundaries provide a compact description of anatomical coverage without first running a segmentation pipeline. CT boundary errors below one vertebral level were promising; larger MR errors likely reflect greater sequence heterogeneity and more examinations without usable vertebral labels.

Compared with prior body-statistics studies, the proposed models cover more outputs and both modalities in one open-source interface. Scout, localizer, dose-metric, and tissue-segmentation approaches can estimate weight or height accurately but depend on particular acquisitions, metadata, or multistage processing~\citep{ichikawa2021,demircioglu2023,ichikawa2024,schenkl2025}. Multitask prediction of MR acquisition parameters has been reported for breast imaging~\citep{konz2024}; the present work extends the concept to heterogeneous clinical CT and MR fields of view and combines acquisition labels with patient characteristics.

External CT results confirmed that performance depends on anatomical coverage. Full coverage produced balanced performance, abdomen-pelvis images gave the lowest weight error, and thorax-only weight error was highest. These findings argue for returning predicted coverage alongside body-statistic estimates so downstream software can judge whether a prediction is appropriate.

The outputs are intended for metadata plausibility checks, retrospective cohort curation, protocol quality control, and pipeline routing rather than replacement of verified clinical measurements. Weight and height predictions permit automated BMI and BSA calculation, but direct measurement remains necessary when errors could change high-risk dosing or ventilator settings~\citep{hendershot2006,sasko2018}. Predicted age is likewise a consistency measure or potential imaging biomarker, not a forensic estimate~\citep{pickhardt2025,wesp2024}.

\subsection{Limitations}

This study has limitations. First, most ground truths came from DICOM metadata or automated algorithms and may contain missing, estimated, or erroneous values. Kernel codes and noise scores are technical surrogates without universal clinical units. Second, some data was not well represented in the training data (e.g. patients younger than 18 years) and thus the model may not be generalizable to this population. Third, external validation was limited to a relatively small CT cohort; independent MR validation and multicenter testing of the additional targets are needed.

In conclusion, one open-source 3D multitask ensemble per modality can rapidly infer patient, acquisition, coverage, and image-quality characteristics from CT and MR volumes. The models are available in TotalSegmentator (\url{https://github.com/wasserth/TotalSegmentator}) and as a web application (\url{https://compute.totalsegmentator.com/body-stats/}).

\bibliographystyle{plainnat}

\clearpage
\onecolumn
\section*{Supplementary Material}

\subsection*{Image-noise Ground Truth}

Noise labels were calculated before CNN training with an automated, modality-specific procedure. TotalSegmentator masks defined the aorta, skeletal muscle, subcutaneous fat, torso fat, and combined body region; the trachea was additionally analyzed for CT. Masks were eroded by physical distance to reduce partial-volume effects. Spatially distributed 10-mm three-dimensional patches were sampled within each tissue region, and a three-dimensional affine intensity trend was fitted and removed from each patch. Local residual noise was estimated robustly as $1.4826$ times the median absolute deviation of the residuals.

For CT, regional noise remained in image-intensity units. The quality-control summary combined skeletal muscle, subcutaneous fat, and torso fat by taking the median across valid regional estimates, requiring at least two valid regions. The 75th-percentile CT target was the median across regions of each region's 75th-percentile patch-noise estimate.

For MR, absolute intensity is arbitrary and not comparable across examinations. Residual patch noise was therefore divided by the absolute median patch signal after excluding patches below a minimum signal-to-noise ratio of 3. Valid relative-noise patches were pooled across skeletal muscle, subcutaneous fat, and torso fat; at least 10 patches from one or more regions were required. The pooled 75th percentile was multiplied by 100 and then by 1.5 during export to place most MR labels on a scale similar to CT. Higher scores indicate greater local image noise. Missing noise labels were imputed with conservative high values (40 for CT and 50 for MR) during model training.

\subsection*{Additional Label Encoding}

Visible vertebrae were identified from C1 through L5 using precomputed vertebral segmentations. A vertebra was considered present when its segmented volume exceeded 100 voxels; the smallest and largest ordinal indices defined the cranial and caudal boundaries. A value of zero denoted unavailable vertebral information. CT convolution kernels were converted to an ordinal sharpness code with manufacturer-specific rules. MR sequence classes were encoded in appearance-related order as T1, proton density, T2, FLAIR, STIR, T2*, susceptibility-weighted, diffusion-weighted, MR angiography, and other. These encoded categorical targets were optimized jointly with the continuous targets and decoded to the nearest valid class after five-fold averaging.

\end{document}